\documentclass[10pt,twocolumn,letterpaper]{article}

\usepackage{cvpr}
\usepackage{times}
\usepackage{epsfig}
\usepackage{graphicx}
\usepackage{amsmath}
\usepackage{amssymb}
\usepackage{algorithm}
\usepackage{algorithmic}
\usepackage[table,xcdraw]{xcolor}

\usepackage[pagebackref=false,breaklinks=true,letterpaper=true,colorlinks,bookmarks=false]{hyperref}

\cvprfinalcopy 

\def\cvprPaperID{6338} 

\ifcvprfinal\fi
\begin{document}

\title{High-Performance Long-Term Tracking with Meta-Updater}

\author{Kenan Dai$^1$, Yunhua Zhang$^2$, Dong Wang$^{1\thanks{Corresponding Author: Dr. Dong Wang, wdice@dlut.edu.cn}}$, Jianhua Li$^1$, Huchuan Lu$^{1,4}$, Xiaoyun Yang$^3$\\
$^1$\normalsize{School of Information and Communication Engineering, Dalian University of Technology, China}\\
$^2$\normalsize{University of Amsterdam} \quad
$^3$\normalsize{China Science IntelliCloud Technology Co., Ltd} \quad
$^4$\normalsize{Peng Cheng Laboratory}\\
{\tt\small dkn2014@mail.dlut.edu.cn, y.zhang9@uva.nl, wdice@dlut.edu.cn}\\
{\tt\small jianhual@dlut.edu.cn, lhchuan@dlut.edu.cn, xiaoyun.yang@intellicloud.ai}
}

\maketitle
\thispagestyle{empty}

\begin{abstract}
Long-term visual tracking has drawn increasing attention because it is
much closer to practical applications than short-term tracking.
Most top-ranked long-term trackers adopt the offline-trained
Siamese architectures, thus, they cannot benefit from great
progress of short-term trackers with online update.
However, it is quite risky to straightforwardly introduce online-update-based
trackers to solve the long-term problem, due to long-term uncertain and noisy
observations.
In this work, we propose a novel offline-trained \textbf{Meta-Updater}
to address an important but unsolved problem: Is the tracker
ready for updating in the current frame?
The proposed meta-updater can effectively integrate geometric,
discriminative, and appearance cues in a sequential manner, and
then mine the sequential information with a designed cascaded
LSTM module.
Our meta-updater learns a binary output to guide the tracker's update
and can be easily embedded into different trackers.
This work also introduces a long-term tracking framework consisting
of an online local tracker, an online verifier, a SiamRPN-based re-detector,
and our meta-updater.
Numerous experimental results on the VOT2018LT, VOT2019LT, OxUvALT,
TLP, and LaSOT benchmarks show that our tracker performs remarkably
better than other competing algorithms.
Our project is available on the
website: \url{https://github.com/Daikenan/LTMU}.
\end{abstract}

\vspace{-4mm}
\section{Introduction}
The study of visual tracking has begun to shift from short-term tracking to large-scale
long-term tracking, roughly due to two reasons.
First, long-term tracking is much closer to practical applications than short-term
tracking.
The average length of sequences in short-term tracking benchmarks (OTB~\cite{OTB2015},
VOT2018~\cite{VOT2018report}, TC128~\cite{TC128}, to name a few) is often at the second level,
whereas the average frame length in long-term tracking datasets (such as VOT2018LT~\cite{VOT2018report},
VOT2019LT~\cite{VOT2019report}, and OxUvALT~\cite{OxUvA}) is at least at the minute level.
Second, the long-term tracking task additionally requires the tracker having the capability
to handle frequent disappearance and reappearance (i.e., having a strong re-detection capability)\footnote{More
resources about long-term tracking can be found in \url{https://github.com/wangdongdut/Long-term-Visual-Tracking}.}.

\begin{figure}[!t]
\begin{center}
\includegraphics[width=1.0\linewidth]{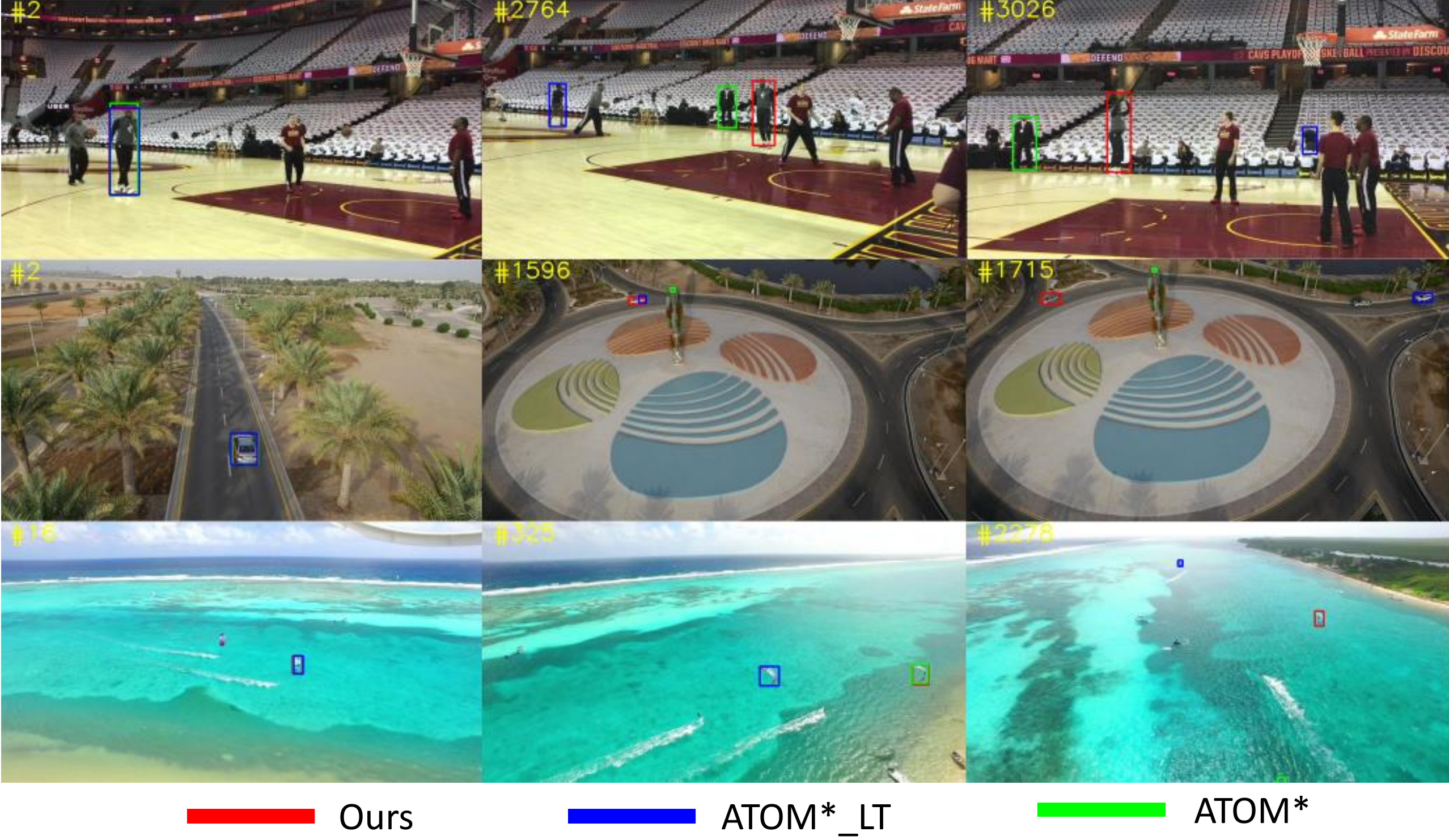}
\end{center}
\vspace{-3mm}
	\footnotesize
\begin{tabular}{cccccc}
\hline
                          & ATOM* &ATOM*\_LT &Ours  &CLGS &SiamDW\_LT\\
\hline
\textbf{F-score}  &0.527 &0.651 &\textcolor{red}{\bf 0.697} &0.674 &0.665\\
\textbf{Pr}  &0.589 &0.685 &0.721 & \textcolor{red}{\bf 0.739} &0.697\\
\textbf{Re}  &0.477  &0.621 &\textcolor{red}{\bf 0.674} &0.619 &0.636\\
\hline
\end{tabular}
	\vspace{1mm}
	\caption{Visualization and comparisons of representative long-term tracking results on VOT2019LT.
	``ATOM*" is our local tracker based on ATOM~\cite{Danelljan-CVPR19-ATOM},
	``Ours" denotes our long-term tracker with meta-update. ``ATOM*\_LT" means ``Ours" without meta-updater.
	``CLGS" and ``SiamDW\_LT" are the second and third best trackers on VOT2019LT. Please see Sections 3 and
	4 for more details.}
	\label{fig-atom-lt}
	\vspace{-4mm}
\end{figure}

Deep-learning-based methods have dominated the short-term tracking field~\cite{DVT-Review,CSA-CVPR2020,DVT-NEW},
from the perspective of either one-shot
learning~\cite{SINT,SiameseFC,SASiam,SiameseRPN,SiamRPNplus,CascadedSiameseRPN,SiameseDW,GradNet}
or online learning~\cite{Nam-CVPR16-MDNet,CCOT,Danelljan-CVPR17-ECO,RTMDNet,DRT,ASRCF,CPF-TIP19,MCPF-TPAMI19,Danelljan-CVPR19-ATOM}.
Usually, the latter methods (e.g., ECO~\cite{Danelljan-CVPR17-ECO}, ATOM~\cite{Danelljan-CVPR19-ATOM})
are more accurate (with less training data) but slower than the former ones (e.g., SiamFC~\cite{SiameseFC},
SiamRPN~\cite{SiameseRPN}).
A curious phenomenon is that few leading long-term trackers exploit online-updated short-term
trackers to conduct local tracking.
MBMD~\cite{Zhang-VOT18-MBMD}, the winner of VOT2018LT, exploits an offline-trained
regression network to directly regress the target's bounding box in a local region, and uses an
online-learned verifier to make the tracker switch between local tracking and global re-detection.
The recent SPLT~\cite{Yan-ICCV19-SPLT} method utilizes the same SiamRPN model
in~\cite{Zhang-VOT18-MBMD} for local tracking.
SiamFC+R~\cite{OxUvA}, the best method in the OxUvALT report, equips the
original SiamFC~\cite{SiameseFC} with a simple re-detection scheme.
An important reason is that online update is a double-edged sword for tracking.
Online update captures appearance variations from both target and background,
but inevitably pollutes the model with noisy samples.
The risk of online update is amplified for long-term tracking, due to long-term uncertain observations.

Motivated by the aforementioned analysis, this work attempts to improve the long-term tracking performance
from two aspects.
First, we design a long-term tracking framework that exploits an online-updated tracker for local tracking.
As seen in Figure~\ref{fig-atom-lt}, the tracking performance is remarkably improved by extending
ATOM* to a long-term tracker (ATOM*\_LT), but it remains worse than the CLGS and SiamDW\_LT
methods.
Second, we propose a novel meta-updater to effectively guide the tracker's update.
Figure~\ref{fig-atom-lt} shows that after adding our meta-updater, the proposed tracker achieves very
promising tracking results.

Our main contributions can be summarized as follows.
\vspace{-3mm}
\begin{itemize}
\setlength{\itemsep}{0pt}
\setlength{\parsep}{0pt}
\setlength{\parskip}{0pt}
\item \emph{A novel offline-trained meta-updater is proposed to address an important
but unsolved problem: Is the tracker ready for updating in the current frame?
The proposed meta-updater effectively guide the update of the online tracker, not only
facilitating the proposed tracker but also having good generalization ability.}
\item \emph{A long-term tracking framework is introduced on the basis of
a SiamRPN-based re-detector, an online verifier, and an online local tracker with our
meta-updater.  Compared with other methods, our long-term tracking framework
can benefit from the strength of online-updated short-term tracker at low risk.}
\item \emph{Numerous experimental results on the VOT2018LT, VOT2019LT,
OxUvALT, TLP and LaSOT long-term benchmarks show that the proposed
method outperforms the state-of-the-art trackers by a large margin. }
\end{itemize}

\vspace{-4mm}
\section{Related Work}
\vspace{-2mm}
\subsection{Long-term Visual Tracking}
\vspace{-2mm}
Although large-scale long-term tracking benchmarks~\cite{VOT2018report,OxUvA} began to emerge since
2018, researchers have attached importance to the long-term tracking task for a long time
(such as keypoint-based~\cite{MUSTer}, proposal-based~\cite{EBT}, detector-based~\cite{TLD,FCLT},
and other methods).
A classical algorithm is the tracking-learning-detection (TLD) method~\cite{TLD}, which addresses long-term tracking
as a combination of a local tracker (with forward-backward optical flow) and a global re-detector
(with an ensemble of weak classifiers).
Following this idea, many researchers~\cite{LCT,FCLT,OxUvA} attempt to handle the long-term tracking problem
with different local trackers and different global re-detectors.
Among them, the local tracker and global re-detectors can also adopt the same powerful
model~\cite{FCLT,SiamRPNplus,Zhang-VOT18-MBMD,Yan-ICCV19-SPLT}, being equipped with
a re-detection scheme (e.g., random search and sliding window).
A crucial problem of these trackers is how to switch the tracker between the local tracker and the global re-detector.
Usually, they use the outputs of local trackers to conduct self-evaluation, i.e., to determine whether the tracker losses
the target or not.
This manner has a high risk since the outputs of local trackers are not always reliable and unexpectedly mislead
the switcher sometimes.
The MBMD method~\cite{Zhang-VOT18-MBMD}, the winner of VOT2018LT, conducts local and global switching
with an additional online-updated deep classifier.
This tracker exploits a SiamPRN-based network to regress the target in a local search region or every sliding window
when re-detection.
The recent SPLT method~\cite{Yan-ICCV19-SPLT} utilizes the same
SiamPRN in~\cite{Zhang-VOT18-MBMD} for tracking and re-detection, replaces the online
verifier in~\cite{Zhang-VOT18-MBMD} with an offline trained matching network,
and speeds up the tracker by using their proposed skimming module.
A curious phenomenon is that most top-ranked long-term trackers (such as MBMD~\cite{Zhang-VOT18-MBMD},
SPLT~\cite{Yan-ICCV19-SPLT}, and SiamRPN++~\cite{SiamRPNplus}), have not adopted excellent
online-updated trackers (e.g., ECO~\cite{Danelljan-CVPR17-ECO}, ATOM~\cite{Danelljan-CVPR19-ATOM})
to conduct local tracking.
One of the underlying reasons is that the risk of online update is amplified for long-term tracking, caused by
long-term uncertain observations.
In this work, we attempt to address this dilemma by designing a high-performance long-term tracker with a meta-updater.

\vspace{-1mm}
\subsection{Online Update for Visual Tracking}
\vspace{-2mm}
\label{sec2-2}
For visual tracking, online update acts as a vital role to capture appearance variations
from both target and its surrounding background during the tracking process.
Numerous schemes have been designed to achieve this goal by using template update~\cite{KOT,DSiam,GradNet},
incremental subspace learning~\cite{IVT,OSPT},
online learning classifiers~\cite{KCF,Nam-CVPR16-MDNet,Danelljan-CVPR17-ECO,Danelljan-CVPR19-ATOM},
to name a few.
However, online update is a double-edged sword in balancing the dynamical information
description and unexpected noise introduction.
Accumulating errors over a long time, collecting inappropriate samples or over-fitting to available
data when the target disappears can easily degrade the tracker and lead to tracking drift, especially
for long-term tracking.
To deal with this dilemma, many efforts have been done at least from two aspects.
The first one aims to distill the online collected samples by recovering or clustering noisy observations~\cite{OSPT,Danelljan-CVPR17-ECO}.
Another effective attempt is to design some criteria for evaluating the reliability of the current tracking
result, to remove the unreliable samples or reject the inappropriate update.
These criteria include the confidence score~\cite{Nam-CVPR16-MDNet},
the maximum (MAX) response~\cite{Danelljan-CVPR19-ATOM}, peak-to-sidelobe rate
(PSR)~\cite{Danelljan-CVPR19-ATOM},
average peak-to-correlation energy~\cite{LMCF}, and MAX-PSR~\cite{FCLT}.
These methods usually utilize the tracker's output to self-evaluate this reliability. But the self-evaluation
of the trackers' reliability with its outputs has inevitable risks, especially when the tracker experiences
the long-term uncertain and noisy observations.
In this work, we propose a novel offline-trained meta-updater to integrate multiple cues in a sequential manner.
The meta-updater outputs a binary score to indicate whether the tracker should be updated or not in the current frame,
which not only remarkably improves the performance of our long-term tracker but also is easy to be embedded
into other online-updated trackers.
Recently, some meta-learning-based methods~\cite{Lee-ECCVW18-MMLT,Park-ECCV18-Meta-tracker,Li-TIP19-RML,Huang-AAAI2019,Park-ICCV19-DML,GradNet} have been presented.
All these methods focus on addressing the ``how to update" problem (i.e., efficiently and/or effectively
updating the trackers' appearance models).
By contrast, our meta-updater is designed to deal with the ``when to update" problem, and it can be
combined with many ``how to update" algorithms to further improve the tracking performance.

\begin{figure}[h]
	\begin{center}
		\includegraphics[width=1.0\linewidth]{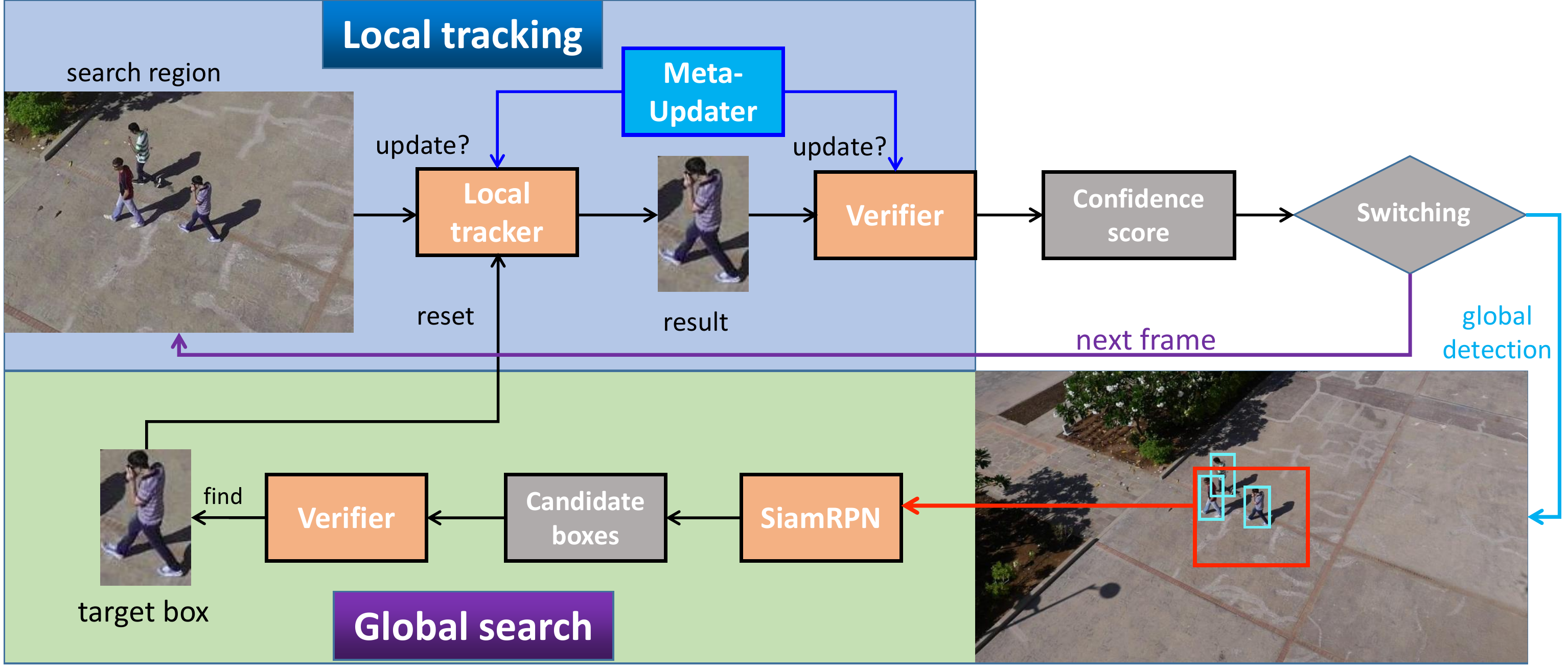}
	\end{center}
	\vspace{-4mm}
	\caption{Proposed long-term tracking framework. Better viewed in color with zoom-in.}
	\label{fig:framework}
	\vspace{-6mm}
\end{figure}
\section{Long-term Tracking with Meta-Updater}
\vspace{-2mm}
\subsection{Long-term Tracking Framework}
\vspace{-2mm}
The overall framework is presented in Figure~\ref{fig:framework}.
In each frame, the local tracker takes the local search region as input, and outputs
the bounding box of the tracked object.
Then, the verifier evaluates the correctness of the current tracking result.
If the output verification score is larger than a predefined threshold, the tracker will continue to
conduct local tracking in the next frame.
If the score is smaller than the threshold, we use the faster R-CNN detector~\cite{mmdetection} to
detect all possible candidates in the next frame and crop the local search region regarding each candidate.
Then, a SiamPRN model~\cite{Zhang-VOT18-MBMD} takes each region as input and outputs corresponding candidate boxes.
These bounding boxes are sent to the verifier for identifying whether there exists the target or not.
When the verifier finds the target, the local tracker will be reset to adapt to the current target appearance.
Before entering into the next frame, all historic information is collected and sent into the proposed
meta-updater.
Finally, the meta-updater guides the online trackers' update.

In this work, we implement an improved ATOM tracker (denoted as ATOM$^*$)
as our local tracker, which applies the classification branch of the
ATOM method~\cite{Danelljan-CVPR19-ATOM}  for localization and exploits
the SiamMask method~\cite{Wang-CVPR19-SiamMask} for scale
estimation\footnote{In the original ATOM method~\cite{Danelljan-CVPR19-ATOM},
the scale estimation is conducted via an offline trained instance-aware
IoUNet~\cite{Jiang-ECCV18-IoUNet}. In practice, we have found the SiamMask
method~\cite{Wang-CVPR19-SiamMask} can provide a more accurate scale
estimation partly due to the strong supervision of pixel-wise annotations.}.
We use the RTMDNet method~\cite{RTMDNet} as our verifier, and its verification
threshold is set to 0.

\noindent \textbf{Strength and Imperfection.} Compared with recent top-ranked
long-term trackers (such as MBMD~\cite{Zhang-VOT18-MBMD} and
SPLT~\cite{Yan-ICCV19-SPLT}), the major strength of our framework lies in embedding
an online-updated local tracker into the long-term tracking framework. This idea makes
the long-term tracking solution benefit from the progress of short-term trackers,
and unifies the short-term and long-term tracking problems as much as possible.
One imperfection is that the risk of online update is amplified due to the long-term
uncertain observations (since the results of any frame except for the first one have no
absolute accuracy during tracking).
Thus, we propose a novel \textbf{Meta-Updater} to handle this problem and
obtain more robust tracking performance.
\vspace{-2mm}
\subsection{Meta-Updater}
\vspace{-2mm}
It is essential to update the tracker for capturing appearance variations from both
target and its surrounding background.
However, the inappropriate update will inevitably make the tracker degrade and
cause tracking drift.
To address this dilemma, we attempt to answer an important but unsolved question:
\textbf{\emph{Is the tracker ready for updating in the current frame?}}
To be specific, we propose a \textbf{Meta-Updater} to determine whether the tracker
should be updated or not in the present moment, by integrating historical tracking results.
These historical results include geometric, discriminative, and appearance cues in a
sequential manner.
We introduce our meta-updater on the basis of an online tracker outputting a
response map in each frame (e.g., ECO~\cite{Danelljan-CVPR17-ECO},
ATOM~\cite{Danelljan-CVPR19-ATOM}).
It is easy to generalize our meta-updater for other types of trackers
(such as MDNet~\cite{Nam-CVPR16-MDNet}).

\vspace{-5mm}
\subsubsection{Sequential Information for Meta-Updater}
\label{sec-3-2-1}
\begin{figure}[htbp]
	\begin{center}
		\includegraphics[width=0.75\linewidth]{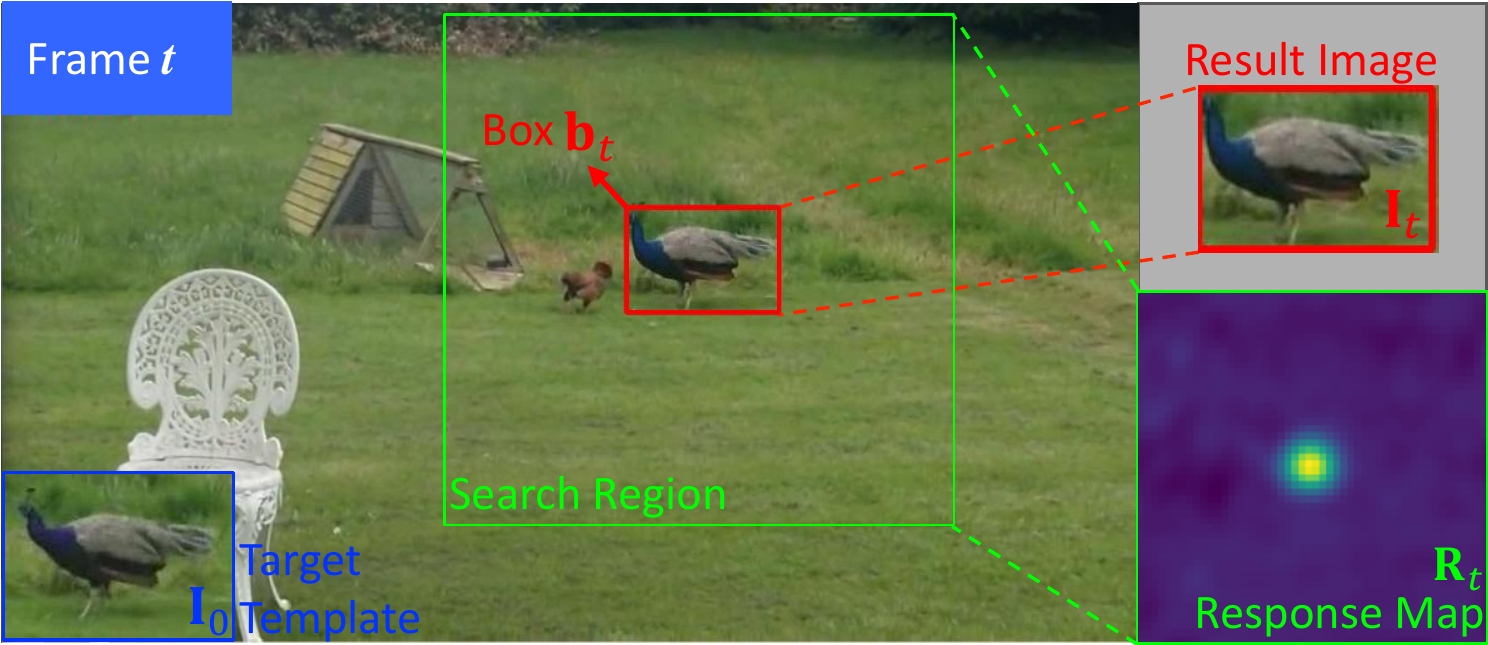}
	\end{center}
	\vspace{-3mm}
	\caption{Intuitive explanations of some notions in this work.}
	\label{fig-intuitive}
	\vspace{-3mm}
\end{figure}

\vspace{-2mm}
Given an online tracker $\mathcal{T}$, in the $t$-th frame, we denote the output
response map as $\mathbf{R}_t$, the output bounding box as $\mathbf{b}_t$, and
the result image (cropped according to $\mathbf{b}_t$) as $\mathbf{I}_t$, respectively.
The target template in the first frame is denoted as $\mathbf{I}_0$.
An intuitive explanation is illustrated in Figure~\ref{fig-intuitive}.
\vspace{-2mm}

We develop our meta-updater by mining the sequential information, integrating geometric,
discriminative, and appearance cues within a given time slice.

\noindent \textbf{Geometric Cue.} In the $t$-th frame, the tracker outputs a bounding
box ${{\bf{b}}_t} = \left[ {{x_t},{y_t},{w_t},{h_t}} \right]$ as the tracking state, where
$\left( {x,y} \right)$ denote the horizontal and vertical coordinates of the up-left corner
and $\left( {w,h} \right)$ are the width and height of the target.
This bounding box itself merely reflects the geometric shape of the tracked object in the
current frame.
However, a series of bounding boxes from consecutive frames contain the important motion
information regarding the target, such as velocity, acceleration, and scale change.

\noindent \textbf{Discriminative Cue.} Visual tracking can be considered as a classification
task to distinguish the target from its surrounding background, thus, an online tracker should
have good discriminative ability itself. We define a confidence score $s_t^C$
as the maximum value of the response map ${{\bf{R}}_t}$ (\ref{eq-maxR}).
For some trackers that do not output any response map (e.g., MDNet~\cite{Nam-CVPR16-MDNet}),
it is also not difficult to obtain this confidence score based on the classification probability or margin.
\begin{equation}
s_t^C = \max \left( {{{\bf{R}}_t}} \right).
\label{eq-maxR}
\end{equation}

\begin{figure}[t]
	\begin{center}
		\includegraphics[width=0.95\linewidth]{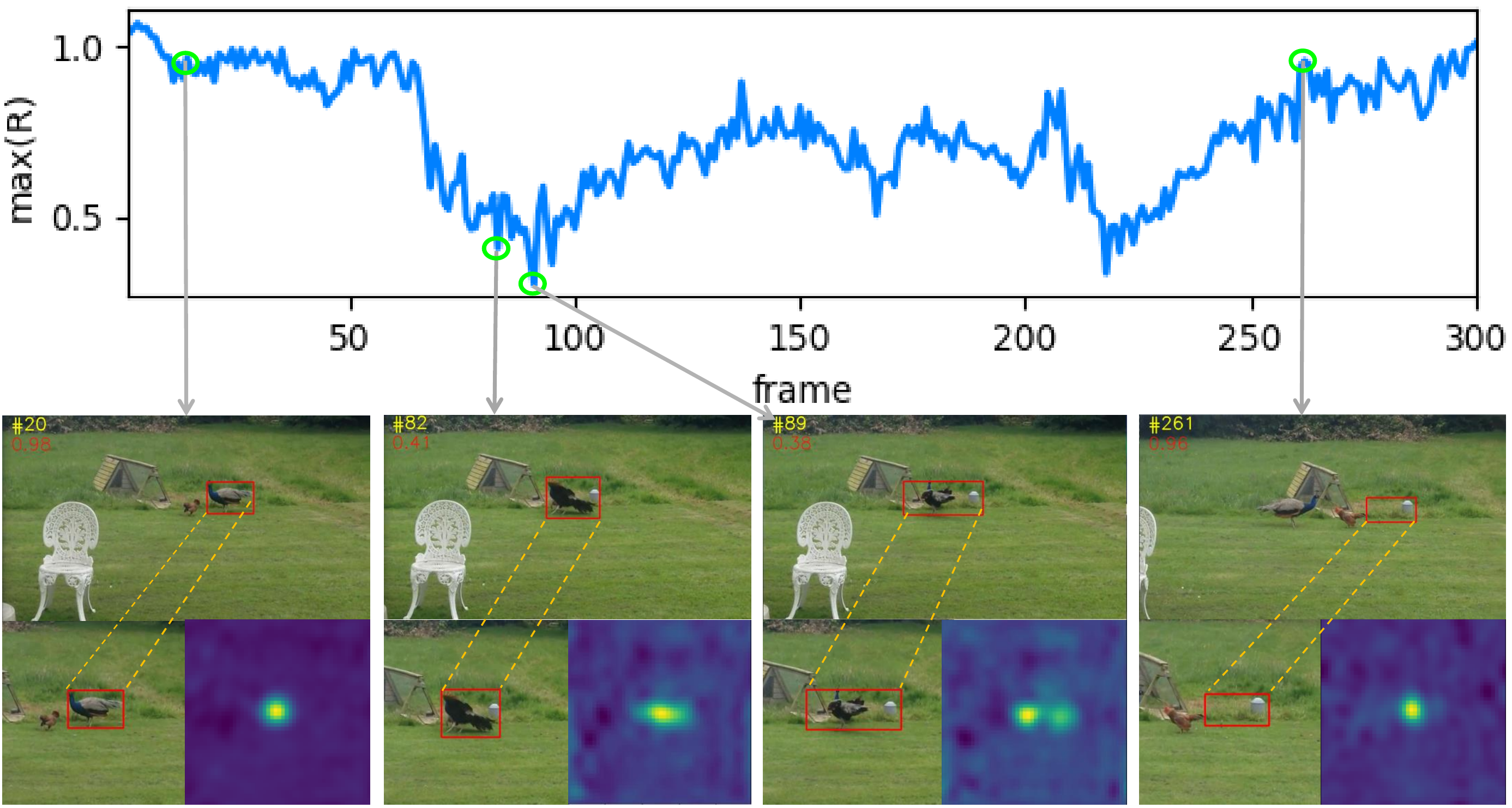}
	\end{center}
	\vspace{-3mm}
	\caption{Illustration of varied confidence scores with representative frames.
	Better viewed in color with zoom-in.}
	\label{fig-response}
	\vspace{-4mm}
\end{figure}

Figure~\ref{fig-response} indicates that the confidence score is not stable during the tracking
process (see $89$-and $261$-th frames).
In this work, we also exploit a convolutional neural network (CNN) to thoroughly mine the
information within the response map, and obtain a response vector $\mathbf{v}_t^R$ as
\begin{equation}
\mathbf{v}_t^R = {f^R}\left( {{{\bf{R}}_t};{{\bf{W}}^R}} \right),
\end{equation}
where ${f^R}\left( {.;.} \right)$ denotes the CNN model with the parameter ${{{\bf{W}}^R}}$.
The output vector $\mathbf{v}_t^R$ implicitly encodes the reliability information of the tracker in
the current frame, and is further processed by the subsequent model.

\noindent \textbf{Appearance Cue.} The self-evaluation of the trackers' reliability with its
outputs has inevitable risks, since online updating with noisy samples often makes the response not
sensitive to appearance variations.
Thus, we resort to a template matching method as a vital supplement, and define an appearance
score as
\begin{equation}
s_t^A = {\left\| {{f^A}\left( {{{\bf{I}}_t},{{\bf{W}}^A}} \right) -
{f^A}\left( {{{\bf{I}}_0},{{\bf{W}}^A}} \right)} \right\|_2},
\end{equation}
where ${f^A}\left( {.,{{\bf{W}}^A}} \right)$ is the embedding function to embed the target
and candidates into a discriminative Euclidean space, ${{\bf{W}}^A}$ stands for its offline
trained network parameters.
As presented in~\cite{Luo2019}, the network ${f^A}\left( {.,{{\bf{W}}^A}} \right)$
can be effectively trained with the combination of triplet and classification loss functions.
The score $s_t^A $ measures the distance between the tracked result ${\bf{I}}_t$ and target
template ${\bf{I}}_0$.
This template matching scheme is not affected by noisy observations.

\noindent \textbf{Sequential Information.}
We integrate the aforementioned geometric, discriminative and appearance cues into
a sequential matrix as
${{\bf{X}}_t} = [{{\bf{x}}_{t - {t_s} + 1}};...;{{\bf{x}}_{t - 1}};{{\bf{x}}_t}] \in
{\mathbb{R}^{d \times {t_s} }}$, where ${{\bf{x}}_t} \in {\mathbb{R}^{d \times {1}}}$
is a column vector concentrated by $s_t^C$, $\mathbf{v}_t^R$, $s_t^A$, and ${{\bf{b}}_t}$.
$d$ is the dimension of concentrated cues, and $t_s$ is a time step to balance the
historical experience and current observation.
This sequential information is further mined with the following cascaded LSTM scheme.
\vspace{-4mm}
\subsubsection{Cascaded LSTM}
\vspace{-2mm}
\noindent \textbf{LSTM.} Here, we briefly introduce the basic ideas and notions of
LSTM~\cite{LSTM} to make this paper self-contained.
Its mathematical descriptions are presented as follows.\\
$\left\{ \begin{array}{l}
 {{\bf{f}}_t} = \sigma \left( {{{\bf{W}}_f}{{\bf{x}}_t} + {{\bf{U}}_f}{{\bf{h}}_{t - 1}} + {{\bf{b}}_f}} \right) \\
 {{\bf{i}}_t} = \sigma \left( {{{\bf{W}}_i}{{\bf{x}}_t} + {{\bf{U}}_i}{{\bf{h}}_{t - 1}} + {{\bf{b}}_i}} \right) \\
 {{\bf{o}}_t} = \sigma \left( {{{\bf{W}}_o}{{\bf{x}}_t} + {{\bf{U}}_o}{{\bf{h}}_{t - 1}} + {{\bf{b}}_o}} \right) \\
 {{\bf{c}}_t} = {{\bf{f}}_t} \odot {{\bf{c}}_{t - 1}} + {{\bf{i}}_t} \odot tanh\left( {{{\bf{W}}_c}{{\bf{x}}_t} + {{\bf{U}}_c}{{\bf{h}}_{t - 1}} + {{\bf{b}}_c}} \right) \\
 {{\bf{h}}_t} = {{\bf{o}}_t} \odot tanh\left( {{{\bf{c}}_t}} \right) \\
 \end{array} \right., $
where $\sigma \left( . \right)$ denotes the element-wise sigmoid function,
$tanh\left( . \right)$ stands for the element-wise tangent operation,
and $\odot$ is the element-wise multiplication.
$\mathbf{W}$, $\mathbf{U}$, and $\mathbf{b}$ denote the weight matrices and bias vector
requiring to be learned. The subscripts $f$, $i$, $o$, and $c$ stand for the forget gate, input
gate, output gate, and memory cell, respectively.
Other variables are defined as follows.
(a) ${{{\bf{x}}_t}}$: the input vector to the LSTM unit;
(b) ${{\bf{f}}_t}$: the forget gate's activation vector;
(c) ${{\bf{i}}_t}$: the input gate's activation vector;
(d) ${{\bf{o}}_t}$: the output gate's activation vector;
(e) ${{\bf{h}}_t}$: the hidden state vector;
and (f) ${{\bf{h}}_t}$: the cell state vector.

\begin{figure}[!t]
	\begin{center}
		\includegraphics[width=1.0\linewidth]{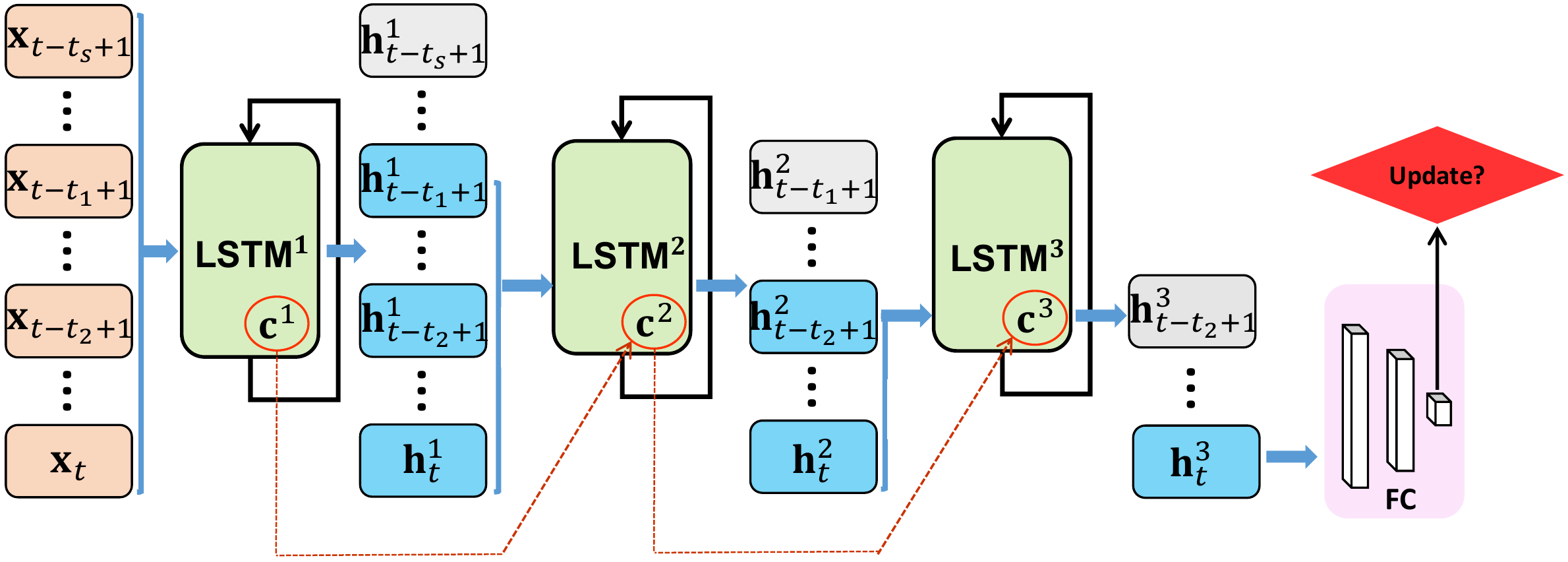}
	\end{center}
	\vspace{-4mm}
	\caption{Proposed three-stage cascaded LSTM.}
	\label{fig-lstm}
	\vspace{-4mm}
\end{figure}

\noindent \textbf{Three-stage Cascaded LSTM.} After obtaining the sequential features
${{\bf{X}}_t}$, presented in Section~\ref{sec-3-2-1}, we feed it into a three-stage
cascaded LSTM model, shown in Figure~\ref{fig-lstm}.
The time steps of three LSTMs gradually decrease to distill the sequential information
and focus on the recent frames.
The input-output relations are presented in Table~\ref{tab:lstminout}. The superscript $i$
denotes the $i$-th stage LSTM.

Finally, the output ${\bf{h}}_t^3$ is processed by two fully connected layers to
generate a binary classification score, indicating whether the tracker should be
updated or not.

\begin{table}[t]
\caption{Input-output relations of our cascaded LSTM model. }
\label{tab:lstminout}
\small
\begin{tabular}{|c|l|}
\hline
Input             & ${{\bf{x}}_{t - {t_s} + 1}},...,{{\bf{x}}_{t - {t_1} + 1}},...,{{\bf{x}}_{t - {t_2} + 1}},...,{{\bf{x}}_t}$\\
\hline
LSTM$^1$ $\to$ LSTM$^2$  & ${\bf{h}}_{t - {t_1} + 1}^1,...,{\bf{h}}_{t - {t_2} + 1}^1,...,{\bf{h}}_t^1;{\bf{c}}_t^1$\\
\hline
LSTM$^2$ $\to$ LSTM$^3$  & ${\bf{h}}_{t - {t_2} + 1}^2,...,{\bf{h}}_t^2;{\bf{c}}_t^2$\\
\hline
Output          &${\bf{h}}_t^3$\\
\hline
\end{tabular}
\vspace{-6mm}
\end{table}

\begin{figure*}[htbp]
	\begin{center}
		\includegraphics[width=0.92\linewidth]{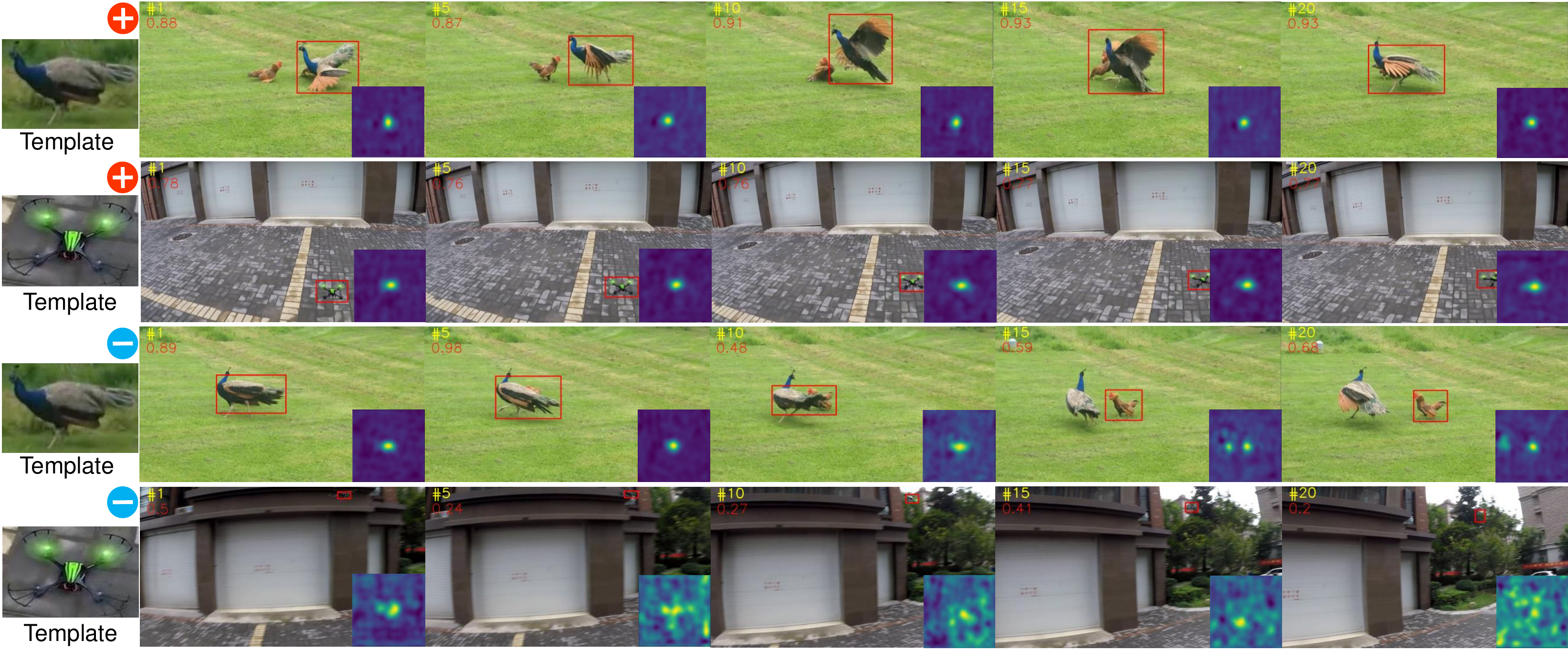}
	\end{center}
	\vspace{-4mm}
	\caption{Illustration of positive and negative samples for meta-updater training. The first two rows illustrate
	two positive examples, whereas the last two rows display the negative ones. In fact,  there is no interval among
	frames, the interval $5$ is merely for clear visualization. }
	\label{fig-pnsample}
	\vspace{-5mm}
\end{figure*}

\vspace{-5mm}
\subsubsection{Meta-Updater Training}
\vspace{-2mm}
\noindent \textbf{Sample Collection.} We run the local tracker on different training
video sequences\footnote{For each sequence, we initialize the target in the first frame
with the groundtruth, and then track it in the subsequent frames. This strictly follows
the experiment setting of online single object tracking. The tracker is online updated
on its own manner.}, and record the tracking results in all frames.
Then, we divide these results into a series of time slices, denoted as
$\mathcal{Y} = \left. {\left( {\left. {{\bf{Y}}_t^v} \right|_{t = {t_s}}^{{t_v}}}
\right)} \right|_{v = 1}^V$. $v$ is the video index, $V$ is the number of training
sequences, and ${{t_v}}$ is the total frame length of the $v$-th video.
${\bf{Y}}_t^v = \left\{ {{\bf{y}}_{t - {t_s} + 1}^v,{\bf{y}}_{t - {t_s} + 2}^v,...,
{\bf{y}}_{t - 1}^v,{\bf{y}}_t^v} \right\}$, where ${{t_s}}$ denotes the time step.
Each time slice ${{\bf{y}}_t^v}$ includes the bounding box, response map, response
score, and predicted target image in the $t$-th frame, along with the corresponding
target template.
See Section~\ref{sec-3-2-1} for more detailed descriptions\footnote{The meaning
of ${{\bf{y}}_t^v}$ is slightly different with that of $\mathbf{x}_t$ because the
parameters of CNN models are also required to be trained.}.

Then, we determine the label of ${{\bf{Y}}_t^v}$ as
 \vspace{-1mm}
\begin{equation}
 l\left( {{\bf{Y}}_t^v} \right) = \left\{ \begin{array}{l}
 1,\;if\;{\bf{IoU}}\left( {{\bf{b}}_t^v,{\bf{g}}_t^v} \right) > 0.5 \\
 0,\;if\;{\bf{IoU}}\left( {{\bf{b}}_t^v,{\bf{g}}_t^v} \right) = 0 \\
 \end{array} \right.,
 \label{eq:iou}
 \vspace{-1mm}
\end{equation}
where ${\bf{IoU}}$ stands for the Intersection-over-Union criterion.
The slices whose IoUs are between $0$ and $0.5$ have been not adopted in the training
phases to guarantee the training convergence.
${\bf{b}}_t^v$ is the output bounding box in the $t$-th frame in video $v$, and
${\bf{g}}_t^v$ is the corresponding groundtruth\footnote{The training sequences
have annotated groundtruth in every frame.}.
Equation (\textcolor{red}{4}) means that the label of a given time slice is determined based
on whether the target is successfully located or not in the current (i.e., $t$-th) frame.
Figure~\ref{fig-pnsample} visualizes some positive and negative samples
for training our meta-updater.

\vspace{-2mm}
\begin{algorithm}[h]
\caption{Iterative Training Scheme}
\label{alg:Framwork}
\begin{algorithmic}[h]
\FOR{$k=0$; $k<K$; $k++$}
\STATE
Run $\left\{ {\mathcal{T},\mathcal{MU}^{k}\left( \mathcal{T} \right)} \right\}$, and record the tracking results\\
Collect training samples ${\mathcal{Y}^k}$ with their labels ${\mathcal{L}^k}$\\
Train the meta-updater ${{\cal M}{{\cal U}^{k+1}}\left( {\cal T} \right)}$\\
\ENDFOR
\end{algorithmic}
\end{algorithm}
\vspace{-2mm}

\noindent \textbf{Model Training.} In this study, the local tracker and its meta-updater
are tightly-coupled.
The tracker affects the sample collection process for training its meta-updater.
The meta-updater will change the tracker's performance, and further affect sample
collection indirectly.
Thus, we propose an iterative training algorithm, listed in \textbf{Algorithm 1}.
The symbol $\left\{ {\mathcal{T},\mathcal{MU}\left( \mathcal{T} \right)} \right\}$
is used to denote a local tracker equipped with its meta-updater
${\mathcal{MU}\left( \mathcal{T} \right)}$.
${{\cal M}{{\cal U}^k}\left( {\cal T} \right)}$ is the learned meta-updater
after the $k$-th iteration ($k=0$ means no meta-updater).
$K$ is set to $3$ in this work.

\vspace{-4mm}
\subsubsection{Generalization ability}
\vspace{-2mm}
The aforementioned introduction is with respect to the online-updated tracker outputting
a response map.
For the trackers without the response map (e.g., MDNet~\cite{Nam-CVPR16-MDNet},
RTMDNet~\cite{RTMDNet}),  we can simply remove the subnetwork ${f^R}$,
and train the meta-updater with the remaining information.
For some trackers those are online updated with accumulated samples over time (such as
ECO~\cite{Danelljan-CVPR17-ECO}), our meta-updater is able to purify the sample
pool used for updating.
For a given frame, if the output of the meta-updater is 0, then the current tracking results
will not be added into the sample pool (i.e., not used for updating).
If an ensemble of multiple online-updated trackers (such as our long-term trackers, ATOM*
for local tracking and RTMDNet for verification),  we can train only one meta-updater with
the information from all trackers as the input, and then use it to guide all trackers' update.
Section~\ref{sec-ga} shows our meta-updater's generalization ability for different trackers.
\vspace{-2mm}
\subsection{Implementation Details}
\vspace{-2mm}
All networks below are trained using the stochastic gradient decent optimizer,
with the momentum of $0.9$.
The training samples are all from the LaSOT~\cite{LaSOT} training set.

\noindent \textbf{Matching Network ${f^A}$. }
The matching network ${f^A}$ adopts the ResNet-50 architecture and takes $107\times 107$
image patches as inputs.
For each target, we randomly sample bounding boxes around the groundtruth in each frame.
We choose the patches with IoU above $0.7$ as the positive data, and use the boxes with high
confidence scores from the SiamRPN-based network~\cite{Zhang-VOT18-MBMD} but not
belonging to the target as the negative data.
The batch size of the network ${f^A}$ is $16$ and we train it for $60000$ iterations.
The initial learning rate is $10^{-4}$ and divided by $10$ every $200000$ iterations.
The matching network is individually trained and fixed when training the remaining networks
of our meta-updater.

\noindent \textbf{Subnetwork ${f^R}$.}
The input response map is first resized to $50\times 50$, processed by two convolutional layers,
and then followed by a global average pooling layer.
The output is a $1\times1\times8$ vector. This subnetwork is jointly trained with the
cascade LSTMs and the two fully connected layers.

\noindent \textbf{LSTMs with fully connected layers.}
The three-stage cascaded LSTMs have $64$ units in each LSTM cell.
$t_s$, $t_1$ and $t_2$ are set to $20$, $8$ and $3$, respectively.
The forget bias is set to $1.0$.
The outputs are finally sent into two fully connected layers with $64$ hidden units to get the final binary value.
Each training stage of LSTM has a batch size of $16$ and is trained by $100, 000$ iterations with the
learning rate of $10^{-4}$.

\vspace{-2mm}
\section{Experiments}
\vspace{-2mm}
We implement our tracker using Tensorflow on a PC machine with an Intel-i9 CPU (64G RAM)
and a NVIDIA GTX2080Ti GPU (11G memory). The tracking speed is approximatively 13
\emph{fps}.
We evaluate our tracker on five benchmarks: VOT2018LT~\cite{VOT2018report},
VOT2019LT~\cite{VOT2019report}, OxUvALT~\cite{OxUvA},
TLP~\cite{moudgil2018long}, and LaSOT~\cite{LaSOT}.

\vspace{-2mm}
\subsection{Quantitative Evaluation}
\vspace{-4mm}
\begin{table}[htbp]
\caption{Comparisons of our tracker and $15$ state-of-the-art methods on the VOT2018LT
dataset~\cite{VOT2018report}. The best three results are shown in \textcolor{red}{\textbf{red}},
\textcolor{blue}{\textbf{blue}} and \textcolor{green}{\textbf{green}} colors, respectively.
The trackers are ranked from top to bottom according to \textbf{F-score}.}
\label{tab-vot18lt}
\vspace{-2mm}
\small
\begin{center}
\begin{tabular}{cccc}
\hline
\textbf{Tracker} & \textbf{F-score}                      & \textbf{Pr}                           & \textbf{Re}                           \\ \hline
\textbf{LTMU(Ours)}             & {\color[HTML]{FE0000} \textbf{0.690}} & {\color[HTML]{FE0000} \textbf{0.710}} & {\color[HTML]{FE0000} \textbf{0.672}} \\
SiamRPN++        & {\color[HTML]{3166FF} \textbf{0.629}} & {\color[HTML]{3166FF} \textbf{0.649}} & {\color[HTML]{3166FF} \textbf{0.609}} \\
SPLT             & {\color[HTML]{32CB00} \textbf{0.616}} & 0.633                                 & {\color[HTML]{32CB00} \textbf{0.600}} \\
MBMD              & 0.610                                 & 0.634                                 & 0.588                            \\
DaSiam\_LT       & 0.607                                 & {\color[HTML]{000000} 0.627}          & 0.588       \\
MMLT                & 0.546                                 & 0.574                                 & 0.521                            \\
LTSINT              & 0.536                                 & 0.566                                 & 0.510                            \\
SYT                   & 0.509                                 & 0.520                                 & 0.499                             \\
PTAVplus           & 0.481                                 & 0.595                                 & 0.404                            \\
FuCoLoT           & 0.480                                 & 0.539                                 & 0.432                             \\
SiamVGG          & 0.459                                 & 0.552                                 & 0.393                             \\
SLT                    & 0.456                                 & 0.502                                 & 0.417                             \\
SiamFC              & 0.433                                 & {\color[HTML]{32CB00} \textbf{0.636}} & 0.328  \\
SiamFCDet        & 0.401                                 & 0.488                                 & 0.341                              \\
HMMTxD         & 0.335                                 & 0.330                                 & 0.339                               \\
SAPKLTF          & 0.323                                 & 0.348                                 & 0.300                               \\
ASMS                & 0.306                                 & 0.373                                 & 0.259                               \\
\hline
\end{tabular}
\end{center}
\vspace{-5mm}
\end{table}

\noindent \textbf{VOT2018LT.}
We first compare our tracker with other state-of-the-art algorithms on the VOT2018LT
dataset~\cite{VOT2018report}, which contains 35 challenging sequences of diverse objects (e.g., persons,
cars, motorcycles, bicycles and animals) with the total length of 146817 frames.
Each sequence contains on average 12 long-term target disappearances, each lasting on average 40 frames.
%
%
The accuracy evaluation of the VOT2018LT dataset~\cite{VOT2018report} mainly includes tracking
precision (\textbf{Pr}), tracking recall (\textbf{Re}) and tracking \textbf{F-score}.
Different trackers are ranked according to the tracking \textbf{F-score}.
The detailed definitions of \textbf{Pr}, \textbf{Re} and \textbf{F-score} can be found in the VOT2018 challenge
official report~\cite{VOT2018report}.

We compare our tracker with the VOT2018 official trackers and three recent methods (i.e.,
MBMD~\cite{Zhang-VOT18-MBMD}, SiamRPN++~\cite{SiamRPNplus}, and SPLT~\cite{Yan-ICCV19-SPLT})
and report the evaluation results in Table~\ref{tab-vot18lt}.
The results show that the proposed tracker outperforms all other trackers by a very large margin.




\noindent \textbf{VOT2019LT.} The VOT2019LT~\cite{VOT2019report} dataset, containing $50$ videos
with 215294 frames in total, is the most recent long-term tracking dataset.
Each sequence contains on average 10 long-term target disappearances, each lasting on average 52 frames.
Compared with VOT2018LT~\cite{VOT2018report}, VOT2019LT poses more challenges since it introduces
15 more difficult videos and some uncommon targets (e.g., boat, bull, and parachute).
Its evaluation protocol is the same as that in VOT2018LT.
Table~\ref{tab:votlt19tab} shows that our trackers achieves the first place on the VOT2019LT challenge.
\vspace{-2mm}
\begin{table}[h]
\caption{Performance evaluation of our tracker and eight competing algorithms on the VOT2019LT dataset.
The best three results are shown in \textcolor[rgb]{1,0,0}{\textbf{red}}
, \textcolor[rgb]{0,0,1}{\textbf{blue}} and  \textcolor[rgb]{0,1,0}{\textbf{green}} colors, respectively.
The trackers are ranked from top to bottom using the \textbf{F-score} measure.}
\label{tab:votlt19tab}
\vspace{-2mm}
\small
\begin{center}
\begin{tabular}{cccc}
\hline
\textbf{Tracker} & \textbf{F-score}                      & \textbf{Pr}                           & \textbf{Re}                           \\ \hline
\textbf{LTMU(Ours)}
					   & {\color[HTML]{FE0000} \textbf{0.697}}
					   & {\color[HTML]{32CB00} \textbf{0.721}}
					   & {\color[HTML]{FE0000} \textbf{0.674}} \\
CLGS             & {\color[HTML]{3166FF} \textbf{0.674}} & {\color[HTML]{3166FF} \textbf{0.739}}
                       & {\color[HTML]{32CB00} \textbf{0.619}} \\
SiamDW\_LT       & {\color[HTML]{32CB00} \textbf{0.665}} & 0.697
                            & {\color[HTML]{3166FF} \textbf{0.636}} \\
mbdet               & 0.567                                 & 0.609                                 & 0.530                                 \\
SiamRPNsLT   & 0.556                                 & {\color[HTML]{FE0000} \textbf{0.749}} & 0.443                                 \\
Siamfcos-LT    & 0.520                                 & 0.493                                 & 0.549                                 \\
CooSiam          & 0.508                                 & 0.482                                 & 0.537                                 \\
ASINT             & 0.505                                 & 0.517                                 & 0.494                                \\
FuCoLoT          & 0.411                                 & 0.507                                 & 0.346                                \\
\hline
\end{tabular}
\end{center}
\vspace{-4mm}
\end{table}

\noindent \textbf{OxUvALT.} The OxUvA long-term (denoted as OxUvALT) dataset~\cite{OxUvA}
contains 366 object tracks in 337 videos, which are selected from YTBB. Each video in this dataset
lasts for average 2.4 minutes, which is much longer than other commonly used short-term datasets
(such as OTB2015~\cite{OTB2015}).
The targets are sparsely labeled at a frequency of 1 Hz.
The dataset was divided into two disjoint subsets, \emph{dev} and \emph{test}.
In this work, we follow the open challenge in OxUvALT, which means that trackers can use any dataset
except for the YTBB validation set for training and use the OxUvALT \emph{test} subset for testing.
In the OxUvALT dataset, three criteria are adopted to evaluate different trackers, including true positive rate
(\textbf{TPR}), true negative rate (\textbf{TNR}) and maximum geometric mean (\textbf{MaxGM}).
\textbf{TPR} measures the fraction of present objects that are reported present as well
as the location accuracy, and \textbf{TNR} gives the fraction of absent frames that are reported as absent.
\textbf{MaxGM} makes a trade-off between \textbf{TPR} and \textbf{TNR} (i.e., $\mathbf{M a x G M}=$
$\max _{0 \leq p \leq 1} \sqrt{((1-p) \cdot \mathbf{T} \mathbf{P} \mathbf{R})((1-p) \cdot \mathbf{T}
\mathbf{N} \mathbf{R}+p)}$), which is used to rank different trackers.
We compare our tracker with three recent algorithms (MBMD~\cite{Zhang-VOT18-MBMD},
SPLT~\cite{Yan-ICCV19-SPLT} and GlobalTrack~\cite{GlobalTrack}) and ten algorithms reported
in~\cite{OxUvA} (such as LCT~\cite{LCT}, EBT~\cite{EBT}, TLD~\cite{TLD},
ECO-HC~\cite{Danelljan-CVPR17-ECO}, BACF~\cite{BACF}, Staple~\cite{Staple},
MDNet~\cite{Nam-CVPR16-MDNet}, SINT~\cite{SINT}, SiamFC~\cite{SiameseFC},
and SiamFC+R~\cite{OxUvA}).
%
%
%
Table~\ref{tab:oxuva} shows that our tracker performs best in terms of \textbf{MaxGM} and \textbf{TPR}
while maintaining a very competitive \textbf{TNR} value.
%

\begin{figure}[htbp]
\vspace{-3mm}
\begin{center}
\begin{tabular}{c@{}c}
\includegraphics[width=0.485\linewidth,height=0.40\linewidth]{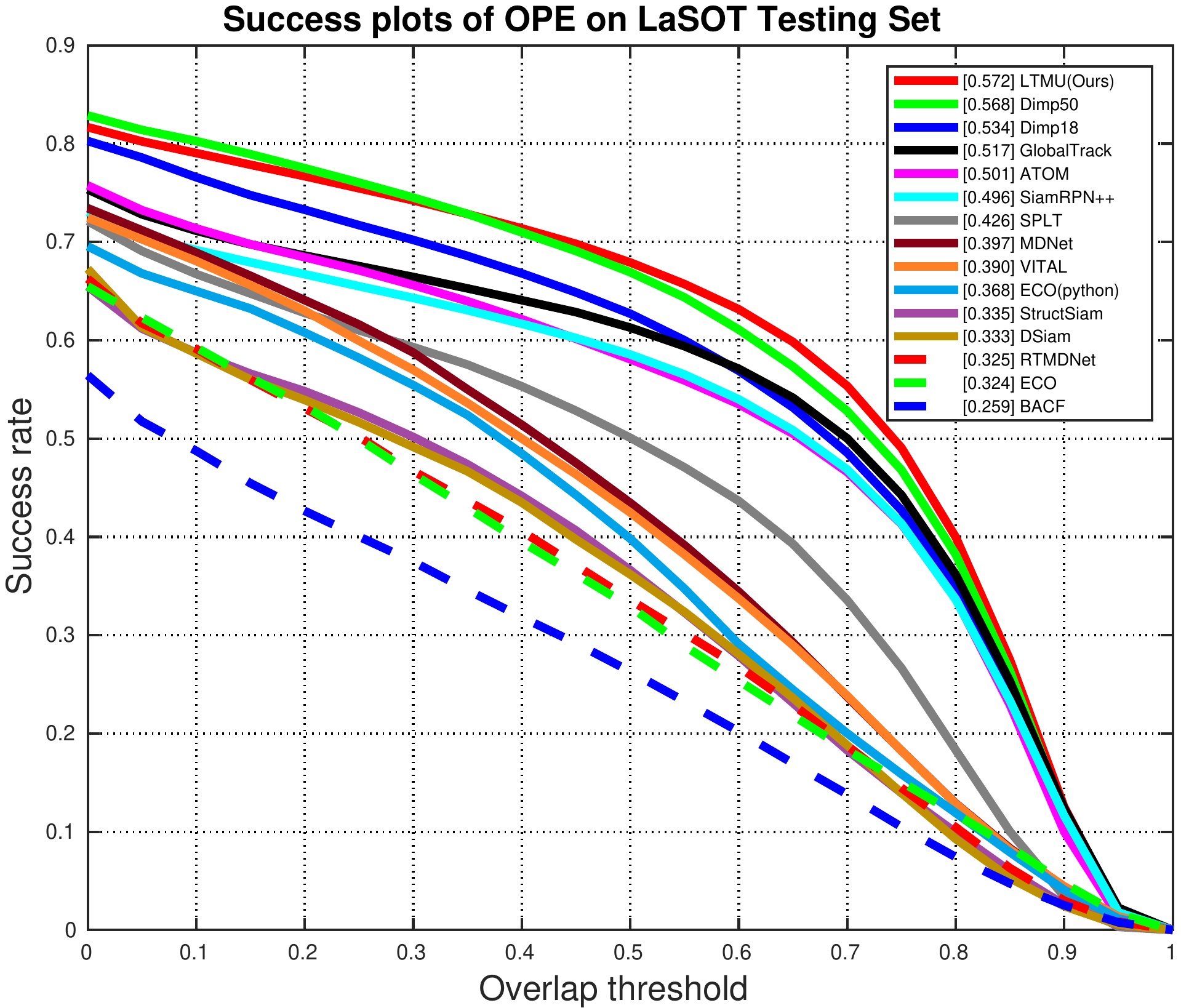} \ &
\includegraphics[width=0.485\linewidth,height=0.40\linewidth]{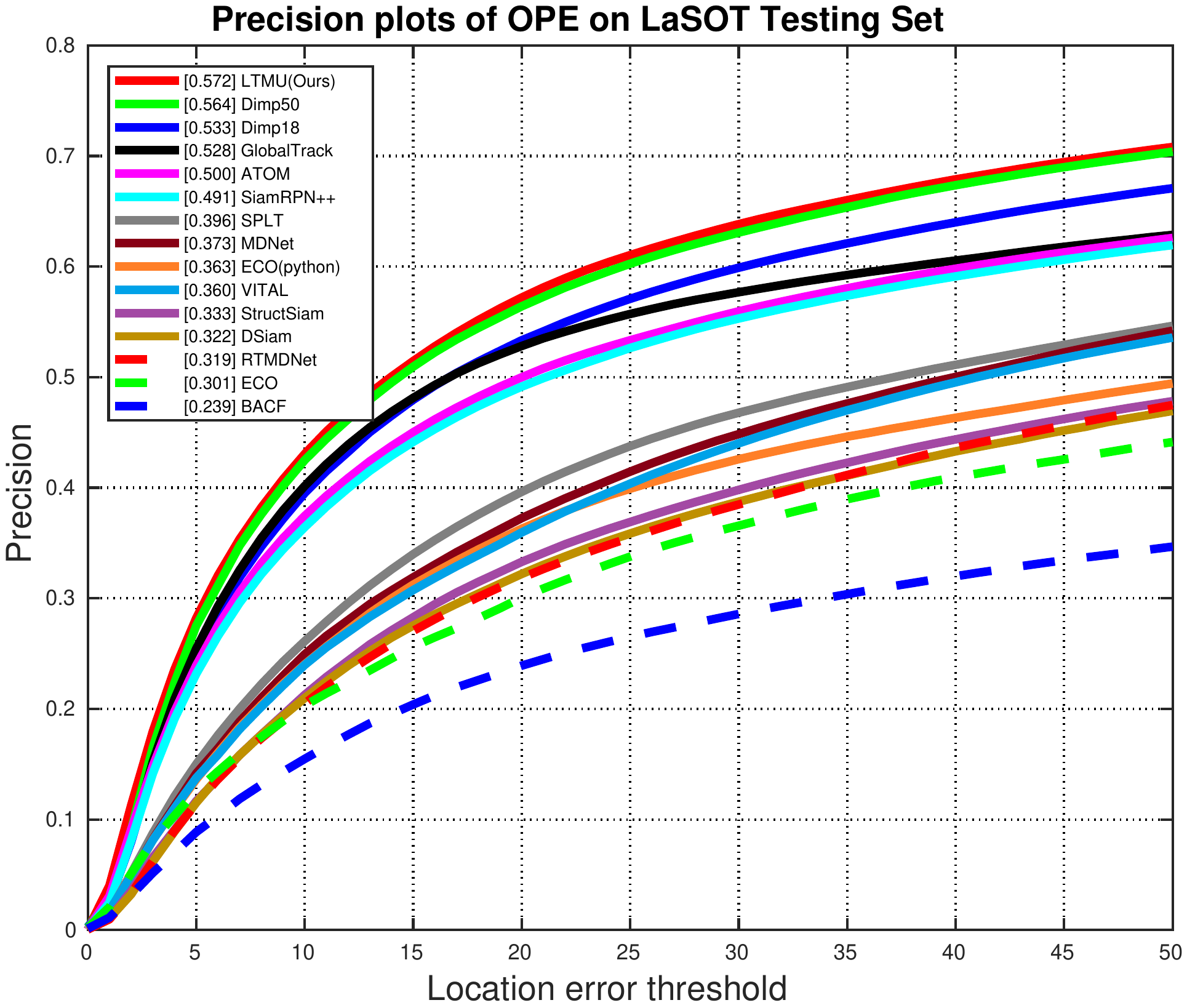} \\
\end{tabular}
\end{center}
\vspace{-5mm}
\caption{One-pass evaluation of different trackers using LaSOT. Better viewed
in color with zoom-in.}
\label{fig-lasotxx}
\vspace{-4mm}
\end{figure}

\vspace{-1mm}
\begin{table}[h]
\caption{Performance evaluation of our tracker and 13 competing algorithms on the OxUvALT dataset.
The best three results are shown in \textcolor{red}{\textbf{red}}, \textcolor{blue}{\textbf{blue}}
and \textcolor{green}{\textbf{green}} colors, respectively. The trackers are ranked from top to bottom
according to the \textbf{MaxGM} values.}
\label{tab:oxuva}
\vspace{-3mm}
\begin{center}
\small
\begin{tabular}{cccc}
\hline
\textbf{Tracker} & \textbf{MaxGM}                        & \textbf{TPR}                          & \textbf{TNR} \\
\hline
\textbf{LTMU(Ours)}    &\textcolor{red}{\textbf{0.751}}  &\textcolor{red}{\textbf{0.749}} &\textcolor{green}{\textbf{0.754}} \\
SPLT                &\textcolor{blue}{\textbf{0.622}} &0.498          &\textcolor{blue}{\textbf{0.776}} \\
GlobalTrack    &\textcolor{green}{\textbf{0.603}} &\textcolor{green}{\textbf{0.574}} & 0.633\\
MBMD            &0.544 &\textcolor{blue}{\textbf{0.609}} &0.485 \\
SiamFC+R      & 0.454                                 & 0.427                                 & 0.481                                 \\
TLD                & 0.431                                 & 0.208                                & \textcolor{red}{\textbf{0.895}} \\
LCT                & 0.396                                 & 0.292                                 & 0.537                                 \\
MDNet            & 0.343                                 & 0.472                                & 0                                     \\
SINT               & 0.326                                 & 0.426                                & 0                                     \\
ECO-HC         & 0.314                                 & 0.395                                & 0                                     \\
SiamFC           & 0.313                                 &0.391                                 & 0                                     \\
EBT                & 0.283                                 & 0.321                                 & 0                                     \\
BACF             & 0.281                                 & 0.316                                 & 0                                     \\
Staple             & 0.261                                 & 0.273                                 & 0                                     \\
\hline
\end{tabular}
\end{center}
\vspace{-7mm}
\end{table}
\noindent \textbf{LaSOT.} The LaSOT dataset~\cite{LaSOT} is one of the most recent large-scale datasets
with high-quality annotations. It contains 1400 challenging sequences (1120 for training and 280 for testing) with
70 tracking categories, with an average of 2500 frames per sequence.
In this work, we follow the one-pass evaluation (success and precision) to evaluate different trackers on the test
set of LaSOT.
Figure~\ref{fig-lasotxx} illustrates both success and precision plots of our tracker and ten state-of-the-art algorithms,
including Dimp50~\cite{Danelljan-ICCV19-DIMP}, Dimp18~\cite{Danelljan-ICCV19-DIMP},
GlobalTrack~\cite{GlobalTrack}, SPLT~\cite{Yan-ICCV19-SPLT},
ATOM~\cite{Danelljan-CVPR19-ATOM}, SiamRPN++~\cite{SiamRPNplus},
ECO(python)~\cite{Danelljan-CVPR17-ECO}, StructSiam~\cite{StructSiam}, DSiam~\cite{DSiam},
and MDNet~\cite{Nam-CVPR16-MDNet}.
Figure~\ref{fig-lasotxx} shows that our tracker achieves the best results among all competing methods.

\begin{figure}[htbp]
\vspace{-3mm}
\begin{center}
\begin{tabular}{c@{}c}
\includegraphics[width=0.485\linewidth,height=0.40\linewidth]{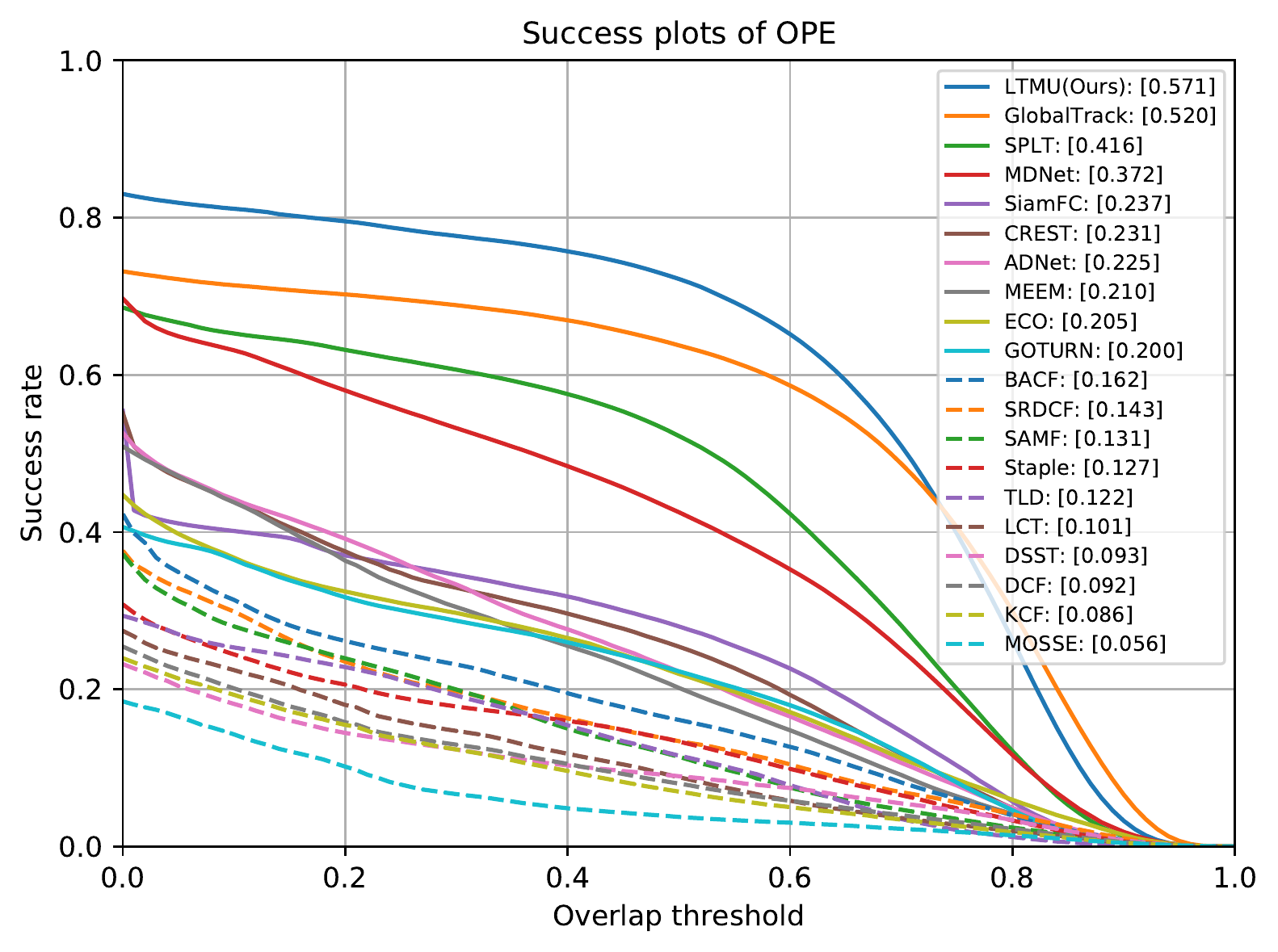} \ &
\includegraphics[width=0.485\linewidth,height=0.40\linewidth]{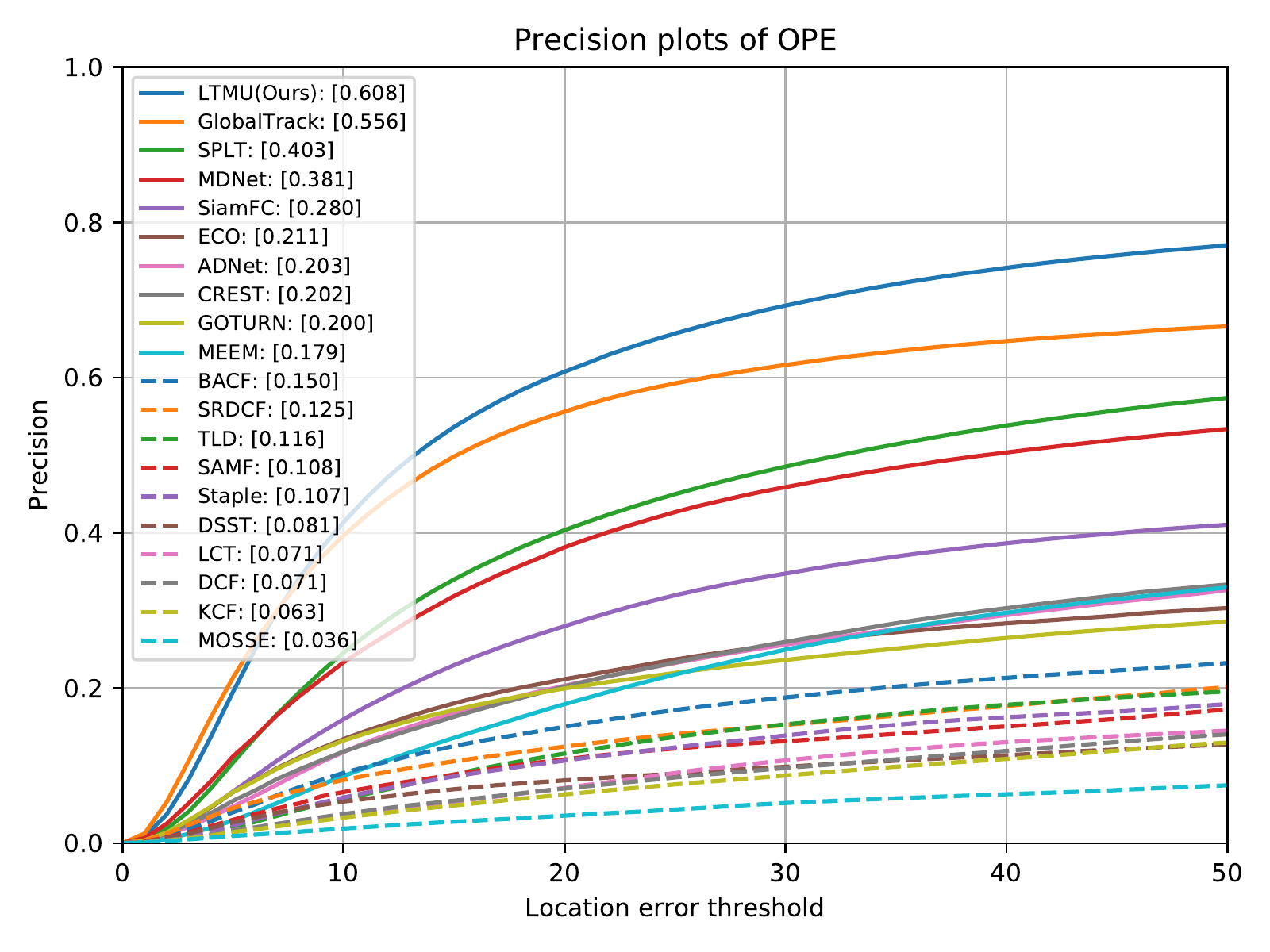} \\
\end{tabular}
\end{center}
\vspace{-5mm}
\caption{One-pass evaluation of different trackers using TLP. Better viewed
in color with zoom-in.}
\label{fig-tlp}
\vspace{-4mm}
\end{figure}

\noindent \textbf{TLP.} The TLP dataset~\cite{moudgil2018long} contains 50 HD videos from
real-world scenarios, with an average of 13500 frames per sequence.
We follow the one-pass evaluation (success and precision) to evaluate different trackers on the
TLP dataset.
As shown in Figure~\ref{fig-tlp}, our tracker achieves the best results among all competing
methods.

\subsection{Ablation Study}
\vspace{-2mm}
In this subsection, we conduct ablation analysis of our meta-updater
using the LaSOT dataset~\cite{LaSOT}.

\noindent \textbf{Different time steps of meta-updater.}
First, we investigate the effects of different time steps.
An appropriate time step could achieve a good trade-off between historical information and
current observations.
Table~\ref{tab:dts} shows that the best performance is obtained when the time step is set to
$20$.

\vspace{-3mm}
\begin{table}[h]
\caption{Effects of different time steps for our meta-updater.}
\vspace{-3mm}
\label{tab:dts}
\small
\begin{center}
\begin{tabular}{cccccc}
\hline
time step  & 5     & 10    & 20   & 30    & 50    \\
\hline
Success      & 0.553 & 0.564 & {\color[HTML]{FE0000} \textbf{0.572}} & 0.570 & 0.567 \\
Precision & 0.548 & 0.561 & {\color[HTML]{FE0000} \textbf{0.572}} & 0.569 & 0.565\\
\hline
\end{tabular}
\end{center}
\vspace{-6mm}
\end{table}

\noindent \textbf{Different inputs for our meta-updater.} For our long-term trackers, the inputs of
the meta-updater include bounding box (B), confidence score (C), response map (R), and appearance
score (A). We verify their contributions by separately removing them from our meta-update.
Detailed results are reported in Table~\ref{tab:updater_input}, showing that each input contributes
to our meta-updater ($w/o$ means `without').

\vspace{-2mm}
\begin{table}[h]
\caption{Effectiveness of different inputs of our meta-updater. }
\vspace{-5mm}
\small
\label{tab:updater_input}
\begin{center}
\begin{tabular}{cccccc}
\hline
different input & $w/o$ C    & $w/o$ R    & $w/o$ B  & $w/o$ A    & Ours                     \\
 \hline
Success         & 0.561 & 0.568 &0.563 & 0.549 & {\color[HTML]{FE0000} \textbf{0.572}} \\
Precision       & 0.558 & 0.566 &0.562 & 0.540 & {\color[HTML]{FE0000} \textbf{0.572}} \\
\hline
\end{tabular}
\end{center}
\vspace{-6mm}
\end{table}

\noindent \textbf{Evaluation of iterative steps.}
Table~\ref{tab:ablationk} shows that the performance is gradually improved with the increase of $k$.

\begin{table}[h]
\vspace{-2mm}
\caption{Evaluation of iterative steps for our cascaded LSTM.}
\vspace{-3mm}
\small
\label{tab:ablationk}
\begin{center}
\begin{tabular}{ccccc}
\hline
$k$         & 0     & 1     & 2    & 3                                     \\
\hline
Success      & 0.539 & 0.562 & {\color[HTML]{000000} 0.568} & {\color[HTML]{FE0000} \textbf{0.572}} \\
Precision    & 0.535 & 0.558 & {\color[HTML]{000000} 0.566} & {\color[HTML]{FE0000} \textbf{0.572}}\\
\hline
\end{tabular}
\end{center}
\vspace{-8mm}
\end{table}

\subsection{Discussions}
\begin{figure}[h]
\vspace{-2mm}
\begin{center}
\begin{tabular}{c@{}c}
\includegraphics[width=0.485\linewidth]{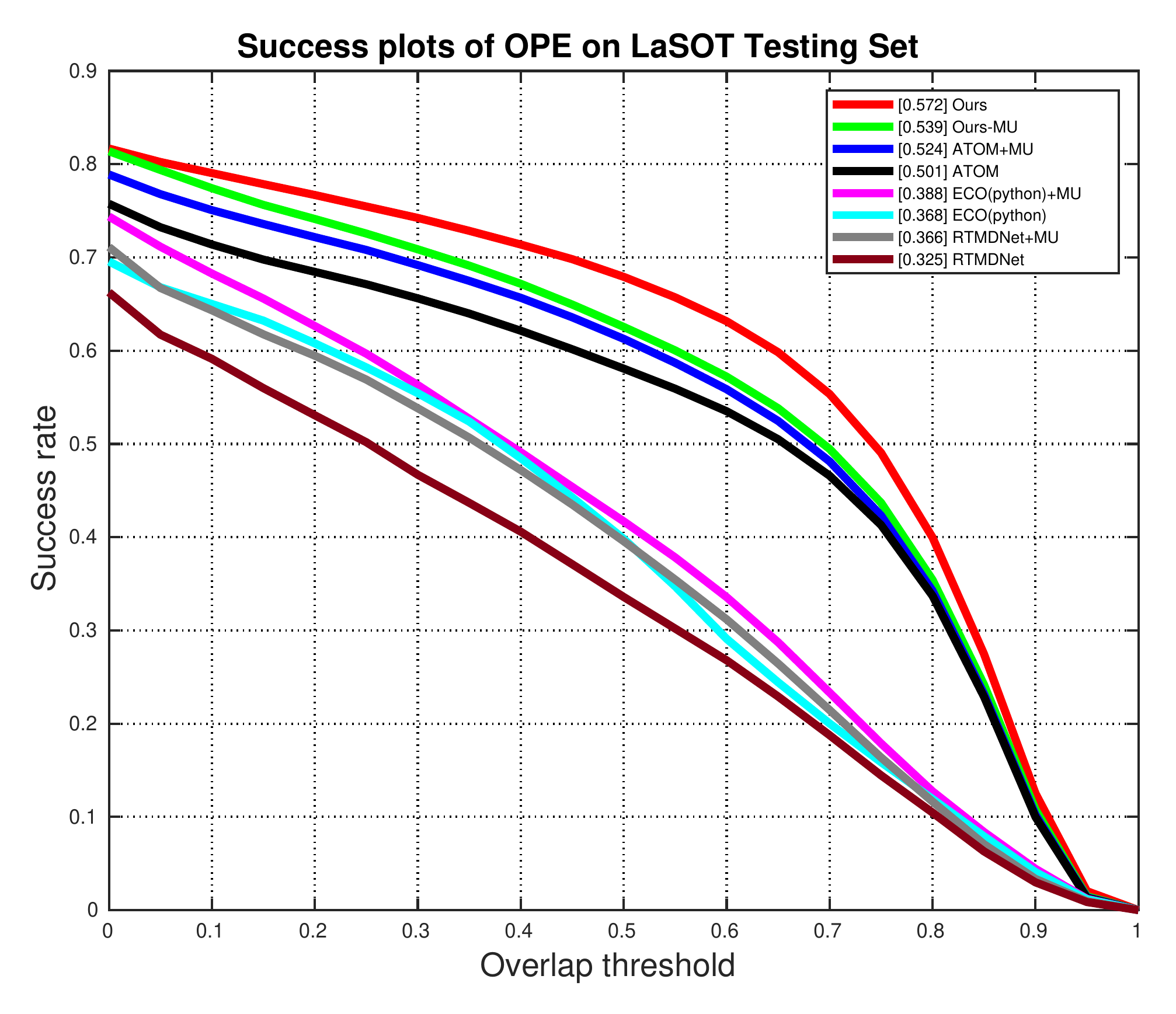}\ &
\includegraphics[width=0.485\linewidth]{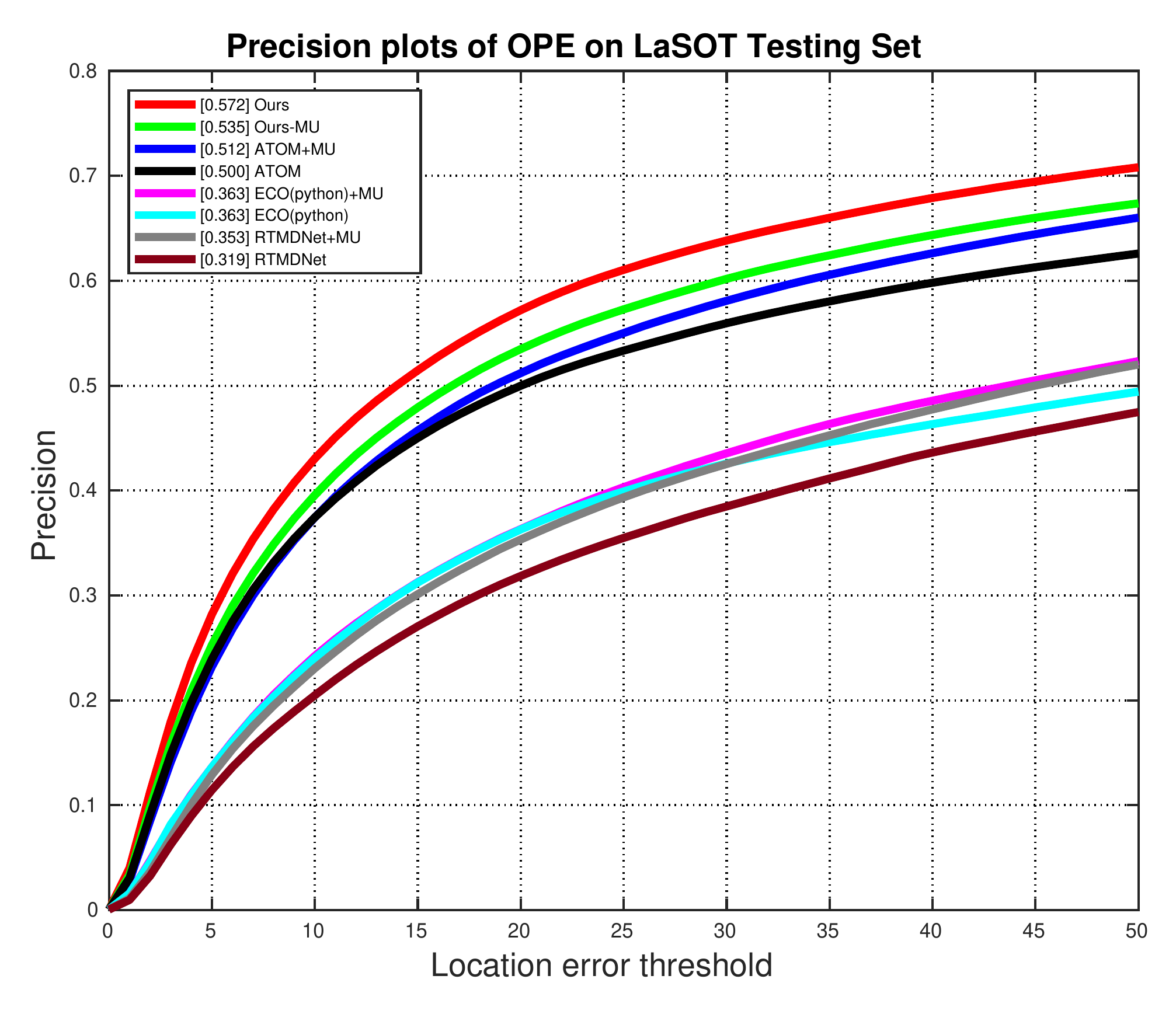}\\
\end{tabular}
\end{center}
\vspace{-5mm}
\caption{Generalization ability of our meta-updater (MU). Different trackers without
and with meta-updater are evaluated using the LaSOT test dataset. Better viewed
in color with zoom-in.}
\label{fig:ablation_lasot}
\vspace{-2mm}
\end{figure}

\vspace{-2mm}
\label{sec-ga}
\noindent \textbf{Generalization ability and speed analysis.} We note that our meta-updater is easy to be embedded
into other trackers with online learning. To show this good generalization ability, we introduce our meta-updater into
four tracking algorithms, including ATOM, ECO (the official python implementation), RTMDNet and our base tracker
(using a threshold to control update).
Figure~\ref{fig:ablation_lasot} shows the tracking performance of different trackers without and with meta-updater
on the LaSOT dataset, and it demonstrates that the proposed meta-updater can consistently improve the tracking
accuracy of different trackers.
Table~\ref{tab:updater_fps} reports the running speeds of those trackers without and with the proposed meta-updater,
which demonstrates that the tracking speeds decrease slightly with an additional meta-updater scheme.
Thus, we can conclude that our meta-updater has a good generalization ability, which can consistently improve
the tracking accuracy almost without sacrificing the efficiency.

\begin{table}[h]
\caption{Speed comparisons of different trackers without and with meta-updater (MU).}
\label{tab:updater_fps}
\footnotesize
\begin{tabular}{ccccc}
\hline
Trackers & ATOM &ECO &RTMDNet  &Ours-MU \\
FPS       & 40   &49  &41 &15 \\
\hline
Trackers &ATOM+MU &ECO+MU  &RTMDNet+MU & Ours\\
FPS      &32 &38 &32 &13\\
\hline
\end{tabular}
\vspace{-1mm}
\end{table}

\noindent \textbf{Why our meta-updater works?}
We run a tracker without and with its meta-updater, and record the trackers' update state
($u = 0, 1$) paired with its ground truth in each frame ($l = 0, 1$).
$u = 1$ means that the tracker has been updated; otherwise, has not been updated.
$l=1$ means that the tracker can be updated; otherwise, cannot be updated.
The definition of ground truth $l$ is the same as equation (\textcolor{red}{4}).
We have the following concepts: (1) true positive (TP): $l = 1, u = 1$;
(2) false positive (FP): $l = 0, u = 1$; (3) true negative (TN): $l = 0, u = 0$; and
(4) false negative (FN): $l = 1, u = 0$.
Then, we can obtain the update precision (Pr), and update recall (Re) as Pr = TP/(TP+FP),
and Re = TP/(TP+FN), respectively.
A higher precision means that the tracker has been updated with less wrong observations.
A higher recall means that the tracker more likely accepts to be updated with correct observations.
We also define a  true negative rate (TNR) to pay much attention to wrong observations
as TNR = TN/(TN+FP).
A higher TNR value means that the tracker rejects to be updated with wrong observations
more strongly.
Table~\ref{tab:update_acc} shows the statistic results of different trackers with and without
their meta-updater modules.
The usage of meta-updater slightly sacrifices the update recall, which means that a portion of
correct observations have not been used to update the tracker in comparison with that without
meta-updater.
This phenomenon affects little on the trackers' performance because correct observations are all for
the same target and have a large amount of redundant information.
In contrast, the usage of meta-updater significantly improves the Pr and TNR values, indicating that
the tracker is much less polluted by wrong observations.
Thus, the risk of online update will be significantly decreased.

\begin{table}[h]
\vspace{-1mm}
\caption{Effectiveness of our meta-updater for different trackers. }
\vspace{-3mm}
\small
\label{tab:update_acc}
\begin{center}
\begin{tabular}{cccc}
\hline
Tracker       			& Pr       & Re             & TNR                 \\ \hline
RTMDNet       		& 0.599  & 0.993        &0.402                            \\
RTMDNet+MU   & 0.909  & 0.902        &0.898                           \\ \hline
ECO           			& 0.583  & 1.000       &0.000                           \\
ECO+MU           & 0.852   & 0.895        &0.803                            \\ \hline
ATOM          	   & 0.765   & 0.997        &0.310                           \\
ATOM+MU        & 0.931   & 0.886        &0.845                           \\ \hline
Ours-MU 		   & 0.867   & 0.994        &0.479                            \\
Ours          		   & 0.952   & 0.874  	    &0.862							  \\ \hline
\end{tabular}
\end{center}
\vspace{-6mm}
\end{table}

\vspace{-3mm}
\section{Conclusions}
\vspace{-2mm}
This work presents a novel long-term tracking framework with the proposed
meta-updater.
Combined with other top-ranked trackers, our framework exploits
an online-update-based tracker to conduct local tracking, which makes the
long-term tracking performance benefit from the excellent short-term
trackers with online update (such as ATOM).
More importantly, a novel meta-updater is proposed by integrating
geometric, discriminative, and appearance cues in a sequential manner
to determine whether the tracker should be updated or not at the present moment.
This method substantially reduces the risk of online update for long-term tracking,
and effectively yet efficiently guides the tracker's update.
Numerous experimental results on five recent long-term benchmarks
demonstrate that our long-term tracker achieves significantly better performance
than other state-of-the-art methods.
The results also indicate that our meta-updater has good
generalization ability.

\noindent {\textbf{Acknowledgement.} The paper is supported in part by
National Natural Science Foundation of China under Grant No. 61872056,
61771088, 61725202, U1903215, in part by the National Key
RD Program of China under Grant No. 2018AAA0102001, and in part by the
Fundamental Research Funds for the Central Universities under Grant No. DUT19GJ201.}

\bibliographystyle{ieee_fullname}
\bibliography{egbib}

\end{document}